\documentclass[11pt,a4paper]{article}
\usepackage[hyperref]{emnlp-ijcnlp-2019}
\usepackage{times}
\usepackage{latexsym}

\usepackage{url}

\usepackage{hyperref}
\usepackage{comment}
\usepackage{bbm}
\usepackage{algorithm}
\usepackage{algpseudocode}
\usepackage{graphicx}
\usepackage{multirow}
\usepackage{booktabs}
\usepackage{color}
\usepackage{subcaption}
\usepackage{amssymb}

\makeatletter
\newcommand{\printfnsymbol}[1]{%
  \textsuperscript{\@fnsymbol{#1}}%
}
\makeatother

\aclfinalcopy 

\title{Language Graph Distillation for Low-Resource Machine Translation}

\author{Tianyu He$^1$ $^{\dagger}$ \thanks{\ \ \ The work was conducted at Microsoft Research Asia.} \ , \  Jiale Chen$^1$ \thanks{\ \ \ Equal contribution} \ , \  Xu Tan$^2$ \printfnsymbol{2}   and  Tao Qin$^2$\\
  $^1$University of Science and Technology of China \\
  $^2$Microsoft Research \\
  {\tt \{hetianyu, chenjlcv\}@mail.ustc.edu.cn} \\  {\tt \{xuta, taoqin\}@microsoft.com} \\}

\date{}

\begin{document}
\maketitle
\begin{abstract}
  Neural machine translation on low-resource language is challenging due to the lack of bilingual sentence pairs. Previous works usually solve the low-resource translation problem with knowledge transfer in a multilingual setting. In this paper, we propose the concept of \textit{Language Graph} and further design a novel graph distillation algorithm that boosts the accuracy of low-resource translations in the graph with forward and backward knowledge distillation. Preliminary experiments on the TED talks multilingual dataset demonstrate the effectiveness of our proposed method. Specifically, we improve the low-resource translation pair by more than $3.13$ points in terms of BLEU score.
\end{abstract}

\section{Introduction}
Neural machine translation (NMT) has witnessed rapid progress in recent years~\citep{DBLP:journals/corr/BahdanauCB14,DBLP:conf/emnlp/LuongPM15,DBLP:conf/nips/SutskeverVL14,DBLP:conf/nips/VaswaniSPUJGKP17,he2018layer,guo2018non}, obtaining good accuracy or even achieving human parity~\cite{DBLP:journals/corr/abs-1803-05567} for rich-resource translation pair. However, there are more than 7000 languages in the world\footnote{https://www.ethnologue.com/browse} and most are low-resource or even zero-resource language pairs, which throw challenges to the data-hungry NMT model. How to improve the translation accuracy with limited bilingual sentence pairs remains a question in NMT. 

Some previous works have studied on this problem, which include: 1) transfer learning~\citep{DBLP:conf/emnlp/ZophYMK16,DBLP:conf/emnlp/GuWCLC18,DBLP:conf/emnlp/NeubigH18} that transfers the knowledge from rich-resource to low-resource languages; 2) pivot translation~\citep{cohn2007machine,utiyama2007comparison,chen2017teacher} that uses a third language to bridge the source to target translation; 3) semi/unsupervised learning~\citep{he2016dual,artetxe2017unsupervised,lample2017unsupervised,DBLP:conf/emnlp/LampleOCDR18} that leverages monolingual data for translation. While these works can improve the accuracy of low-resource translation to some extent, they either just leverage some rich-resource language pairs for knowledge transfer and pivoting, or just leverage the monolingual data of the language itself, without considering the relationship between languages and the monolingual data of each language in a global perspective. 

In this paper, we propose the concept of language graph, where each node and edge in the graph represents the language and translation pair respectively. We further propose a graph distillation algorithm based on language graph which boosts the accuracy of low-resource translation with forward and backward knowledge distillation. We formulate the graph distillation algorithm like this: (1) we choose the edges (translation pairs) in the graph which are of high-potential to improve; (2) for each of the potential edge, we find the high-quality paths that connect the source and target language of this edge; (3) we distill the knowledge from the high-quality paths through the forward and backward translation directions to improve the high-potential edges. We conduct preliminary experiments on the TED talks multilingual dataset which contains translations sentence pairs between more than 50 languages. Our graph distillation algorithm can improve the low-resource language pairs by more than $3.13$ points in terms of BLEU score.

Our contributions are listed as follows. (1) We propose the concept of language graph, which can model the neural machine translation in multilingual setting. (2) We design a novel graph distillation algorithm to improve the low-resource machine translation. (3) Preliminary experiments on the TED talks multilingual dataset demonstrate the effectiveness of our method.

\section{Related Work}


The related works on low-resource machine translation can be classified in three categories. The basic idea of the first category is to transfer the knowledge from rich-resource to low-resource languages~\citep{DBLP:conf/emnlp/ZophYMK16,DBLP:conf/naacl/GuHDL18,DBLP:conf/emnlp/GuWCLC18,DBLP:conf/emnlp/NeubigH18,tan2018multilingual}. The second category mainly leverages a third language as the pivot to enable the translation~\citep{leng2019unsupervised,cohn2007machine,DBLP:conf/acl/WuW07,utiyama2007comparison,DBLP:conf/emnlp/FiratSAYC16,DBLP:journals/tacl/JohnsonSLKWCTVW17,DBLP:journals/corr/HaNW16}, considering there are enough bilingual sentence pairs connected with the pivot language. The last category mainly leverages the monolingual sentences of the low-resource language and formulates it as a semi-supervised or unsupervised problem. ~\citet{he2016dual} proposed dual learning to solve the low-resource translation based on few bilingual but large monolingual sentence pairs. ~\citet{song2019mass,artetxe2017unsupervised,lample2017unsupervised,DBLP:conf/emnlp/LampleOCDR18} leveraged purely unsupervised learning for machine translation.

There are few works on language graph, let alone using language graph for machine translation. ~\citet{ronen2014links} formulated the language network through the connections in book translations, multiple language editions of Wikipedia, and Twitter. ~\citet{samoilenko2016linguistic} studied the network of global interconnections between language communities, based on shared co-editing interests of Wikipedia editors. The above works all concentrate on the analysis of language itself with the help of language network based on other data, such as the co-edit activities in Wikipedia, while we leverage language graph for machine translation which are directly derived from the multilingual translation dataset. The works that are mostly related to but far from ``graph" in machine translation is the pivot translation, where a third language is leveraged to bridge the translation from source to target language.


\section{Language Graph Distillation}
In this section, we first give description about the concept of language graph for machine translation, and then formulate the graph distillation algorithm for low-resource machine translation.

\subsection{Language Graph}
\label{sec:lan_graph}
Denote graph $G=(V, E)$ where $V$ is the set of nodes and $E$ is the set of edges. 
We formulate the language as the node $v \in V$ and translation pair as the edge $e \in E$ in Graph $G$. We will use node and language, edge and translation pair interchangeably. Denote weight $W(e)$ as the translation accuracy of the corresponding translation pair $e$. Therefore $G$ is a direct graph where the weight $w$ are different in two directions between two languages. Denote $D(v)$ as the number of sentence pairs for language $v$, where $D_{b}(v)$ and $D_{m}(v)$ represent the bilingual and monolingual data on language $v$. The number of the bilingual data on a language are the total bilingual data of the language pairs related to this language. Similarly, denote $D(e)$ as the number of bilingual sentence pairs for language pair $e$.

The multilingual machine translation problem on graph $G$ can be formulated as follows: Given a set of languages $V$, translation pairs $E$, bilingual data $D(e)$ for $e \in E$ and monolingual data $D_{m}(v)$ for $v \in V$, the multilingual translation is to develop machine translation algorithm, in order to maximize each $W(e)$ for $e \in E$ or $\Sigma_{e\in E} W(e)$.

\subsection{Graph Distillation Algorithm}
In this subsection, we first describe some concepts used in our graph distillation algorithm, and then formulate detailed steps of this algorithm.

\paragraph{Multi-Hop Translation Path} For source language $v_s$ and target language $v_t$, there exist several forward and backward paths that connect $v_s$ to $v_t$. For example, the one-hop forward path $v_s\to v_t$, two-hop forward path $v_s\to v_1\to v_t$, or two-hop backward path $v_t\to v_2\to v_s$, where $v_1$ and $v_2$ are pivot languages.


\paragraph{Multi-Hop Accuracy Table}
As there are many forward and backward paths between the source and target languages, we maintain translation accuracy tables for the paths with different length of hops between any languages. Denote the accuracy table as $W^h$, where $h \geq 1$ represent the number of hops for each path. $W^h$ is a $h+1$-dimensional matrix, where the first and last dimension represent the source and target language respectively, and each entry in the matrix represents the accuracy for the corresponding $h$-hop path between a source and a target language. For example, $W^h$ is a two-dimensional matrix when $h=1$, where each row and column represent a source and target language respectively, and each entry in the matrix represents the accuracy for an one-hop path.

\paragraph{Forward/Backward Distillation}
For a low-resource translation pair $e=v_s\to v_t$, their direct translation path $v_s\to v_t$ is usually of low translation quality. However, some of the forward paths related to the two languages have rich-resource sentence pairs and the diversity of these paths can provide additional information, which will help improve the direct low-resource translation pair. On the other hand, the paths in the backward translation direction of $e$ are also helpful, as back-translation is useful for neural machine translation~\cite{sennrich2015improving,he2016dual}. But different from back-translation which just leverages the reverse direction, we also leverage the multi-hop backward paths, e.g., $v_t\to v_1\to v_s$, $v_t\to v_2\to v_1\to v_s$. 

Specifically, we use sequence-level knowledge distillation~\citep{DBLP:conf/emnlp/KimR16} to transfer the knowledge from the forward and backward paths to the low-resource translation pairs. The forward and backward paths that are of comparable or better accuracy than the low-resource translation pair will be used to translate the bilingual and monolingual sentences to generate pseudo translation sentence pairs. These generated pseudo sentence pairs are added into the original bilingual sentences of the low-resource pair to boost the accuracy.

\paragraph{Distillation Path Selection}
There are so many paths in the graph, we need be selective to choose which low-resource pair to improve in the current step. For the chosen low-resource pair, we need also choose the related forward and backward paths with good quality for knowledge distillation. We use a greedy strategy to choose the low-resource pair to improve. We define the potential of a language pair as the gap between the accuracy of the direct translation and the multi-hop paths. The more the multi-hop paths are better than the direct translation, the more potential this language pair has. For each chosen translation pair, we then choose the forward and backward paths with the top-K best accuracy respectively for knowledge distillation. For the backward multi-hop paths, we also leverage the one-hop path, which can be considered as standard back-translation.

\begin{algorithm}[t]
\caption{Language Graph Distillation}
\label{alg_graph_iter}
\begin{algorithmic}[1]
\State \textbf{Input}: Graph $G$, which includes a set of languages $V$ and translation pairs $E$, bilingual data $D(e)$ for $e \in E$ and monolingual data $D_{m}(v)$ for $v \in V$. Threshold of accuracy improvement $\tau$, the maximum hop size $H$. 

\State \textbf{Initialize}: Set iteration step $T$ = $0$. Train the multilingual model $\theta_0$ on the available bilingual sentence $D(e)$ for $e \in E$.  Set the accuracy improvement $\sigma$ = $\infty$.
\State \textbf{while} $\sigma > \tau$ \textbf{do}
\State ~~~~ $T$ = $T$+$1 $
\State ~~~~ Construct table $W_{T}^{h}$ for $h \in [H]$. 
\State ~~~~ Select high-potential edges $E_{T}$. 
\State ~~~~ \textbf{for} $e_{T} \in E_{T}$ \textbf{do}
\State ~~~~~~~~ Generate pseudo sentence pairs for $e_{T}$ with forward and backward distillation.
\State ~~~~ \textbf{end for}
\State ~~~~ Train multilingual model $\theta_T$ for $E_{T}$, and get the average accuracy improvements $\sigma$.

\State \textbf{end while}
\end{algorithmic}
\label{alg_graph}
\end{algorithm}

\paragraph{Iteration on the Graph} We conduct the forward and backward distillation iteratively on the graph. For each iteration, we choose the low-resource language pairs with the highest potential currently and the associated forward and backward multi-hop paths, and train the multilingual model for the chosen low-resource pairs with the generated pseudo sentence pairs by the multi-hop paths. After the model is converged, we update the multi-hop accuracy table $W_{T}^h$ for iteration $T$. We repeat the iteration until the accuracy table of the one-hop translation pair is converged. 

The detailed steps for the graph distillation algorithm are shown in Algorithm~\ref{alg_graph}.


\section{Experiments and Results}
\label{ssec:layout}
In this section, we describe the experiment settings and show the preliminary results of our proposed graph distillation algorithm. Note that this work is still in progress.

\subsection{Experiment Setup}
\paragraph{Dataset} We use the common corpus of TED talks which contains bilingual sentences pairs between more than 50 languages~\citep{Ye2018WordEmbeddings}\footnote{https://github.com/neulab/word-embeddings-for-nmt} and also use the monolingual data from TED talks\footnote{https://github.com/ajinkyakulkarni14/TED-Multilingual-Parallel-Corpus/tree/master/Monolingual\_data}. Since we just verify the effectiveness of our framework in this paper, we randomly select $9$ languages to construct the language graph in our experiments for simplicity, as illustrated in Table~\ref{data_pivot_direct_compare}.


\begin{table}[h]
	\small
	\centering
	\begin{tabular}{c|cccccc}
		\toprule
		& Fi & He & Nb & Sk & Sl \\
		\midrule
		Ar & & \checkmark & & \checkmark & \\
		En & \checkmark & \checkmark & \checkmark &  & \checkmark \\
		Fr & \checkmark& & \checkmark & & \\
		Ru & & & &\checkmark & \checkmark \\
\bottomrule
\end{tabular}
\caption{Languages used in our experiments. $\checkmark$ represents there are bilingual data between the language in the row and column. There are bilingual data between any two of Ar, En, Fr and  Ru.} 
\label{data_pivot_direct_compare}
\end{table}

\paragraph{Model Configurations}
We use the Transformer~\citep{DBLP:conf/nips/VaswaniSPUJGKP17} as the basic NMT model structure. The model hidden size, feed-forward size, number of layer is $256$, $1024$ and $6$ respectively. For the basic multilingual model training, we add a special tag to the encoder input to determine which target language to translate, following the practice in~\citet{johnson2017google}. 

\paragraph{Training and Inference}
For the basic multilingual model training, we upsample the data of each language to make all languages have the same size of data. The mini batch size is set to roughly 4096 tokens. We train the models on 4 NVIDIA V100 GPUs. We follow the default parameters of Adam optimizer~\citep{kingma2014adam} and learning rate schedule in~\citet{DBLP:conf/nips/VaswaniSPUJGKP17}. During inference, we decode with beam search and set beam size to $6$ and length penalty $\alpha=1.1$ for all the languages. We evaluate the translation quality by tokenized case sensitive BLEU~\citep{DBLP:conf/acl/PapineniRWZ02} with multi-bleu.pl\footnote{https://github.com/moses-smt/mosesdecoder/blob/ master/scripts/generic/multi-bleu.perl}. 

\subsection{Results}


In this section, we show the preliminary experimental results of the proposed language graph distillation. We select $3$ most potential language pairs in the each iteration and perform the first two iteration steps, and show the results in Table~\ref{results}. The third column is the BLEU score obtained by the basic multilingual model (Initial). The forth, five and sixth columns are the results trained with only one-hop back-translation (+BT), only forward distillation (+Forward) and our language graph distillation (+Graph: both forward and backward distillation) respectively.
The +BT baseline is obtained by training the selected language pairs with one-hop back-translation data, while the +Forward baseline is obtained by training the selected language pairs with all the forward distillation data.

\begin{table}[h]
\small
\centering
\begin{tabular}{c|ccccc}
\toprule
$T$ & Pair & Initial & +BT & +Forward & +Graph \\
\midrule
 & Ar$\to$Fi  & $5.70$ & $6.53$ & $6.67$ & $\textbf{7.58}$ \\
0 & He$\to$Fi & $7.42$ & $8.29$ & $8.61$ & $\textbf{9.04}$ \\
 & Nb$\to$Sl & $8.58$ & $7.35$ & $7.23$ & $\textbf{9.67}$ \\
\hline
Av. & & $7.23$ & +$0.16$ & +$0.27$ & +$\textbf{1.57}$ \\
\hline
 & Ar$\to$Nb & $10.90$ & $12.83$ & $12.89$ & $\textbf{13.92}$ \\
1 & He$\to$Nb & $14.11$ & $\textbf{18.78}$ & $17.93$ & $17.64$ \\
 & Sk$\to$Nb & $13.14$ & $14.73$ & $14.87$ & $\textbf{16.00}$ \\
\hline
Av. & & $12.72$ & +$2.73$ & +$2.51$ & +$\textbf{3.13}$ \\
\bottomrule
\end{tabular}
\caption{The language pairs improved in the first two iterations of our method. The results demonstrate that our method achieves better accuracy than the +BT and +Forward baseline on the low-resource translation edges in the graph. $T$ indicates iteration step. Av. indicates averaged BLEU scores. Note that, we just demonstrate the preliminary results to verify the effectiveness of our language graph distillation method.} 
\label{results}
\end{table}

It can be seen that at each iteration step $T$, our method significantly outperforms all baselines in most cases. 
Note that, for He$\to$Nb translation, our method is slightly worse than the model trained with one-hop back-translation (+BT). The baseline model for He$\to$Nb translation has already achieved good performance, and thus is hard to be further improved by multi-hop forward/backward distillation. However, since the back-translation is the special case of the backward distillation in our method, we can selectively choose each configuration above (+BT, +Forward) to achieve higher BLEU score for each language pair.

\section{Conclusion}
In this paper, we introduced the concept of language graph and further proposed the graph distillation algorithm to boost the accuracy of low-resource machine translation. The preliminary results on the multilingual and low-resource translation dataset demonstrate the effectiveness of our method and show potential for further improvements. For future work, we will work on optimizing the iteration scheme in our algorithm and taking full advantages of more edges (translation pairs) in our language graph distillation.

\bibliography{lan_graph}
\bibliographystyle{acl_natbib}

\end{document}